# A Trustable LSTM-Autoencoder Network for Cyberbullying Detection on Social Media Using Synthetic Data

Mst Shapna Akter[*], Hossain Shahriar[†], Alfredo Cuzzocrea[‡]

[*] Department of Intelligent Systems and Robotics, University of West Florida, USA
[†] Center of Cybersecurity, University of West Florida, USA
[‡] Department of Electrical and Computer Engineering (ECE), North South University, Bangladesh
[§] iDEA Lab, University of Calabria, Rende, Italy

{[*] makter2@students.kennesaw.edu, [†] hshahria@kennesaw.edu, [‡] nova.ahmed@northsouth.edu, [§] alfredo.cuzzocrea@unical.it }

*Abstract*—Social media cyberbullying has a detrimental effect on human life. As online social networking grows daily, the amount of hate speech also increases. Such terrible content can cause depression and actions related to suicide. This paper proposes a trustable LSTM-Autoencoder Network for cyberbullying detection on social media using synthetic data. We have demonstrated a cutting-edge method to address data availability difficulties by producing machine-translated data. However, several languages such as Hindi and Bangla still lack adequate investigations due to a lack of datasets. We carried out experimental identification of aggressive comments on Hindi, Bangla, and English datasets using the proposed model and traditional models, including Long Short-Term Memory (LSTM), Bidirectional Long Short-Term Memory (BiLSTM), LSTM-Autoencoder, Word2vec, Bidirectional Encoder Representations from Transformers (BERT), and Generative Pre-trained Transformer 2 (GPT-2) models. We employed evaluation metrics such as f1-score, accuracy, precision, and recall to assess the models' performance. Our proposed model outperformed all the models on all datasets, achieving the highest accuracy of 95%. Our model achieves state-of-the-art results among all the previous works on the dataset we used in this paper.

## I. INTRODUCTION

The invention of the World Wide Web (WWW) exerted a significant influence on social media, allowing users to share various types of content, including informative, entertaining, and personal information, quickly and easily through digital devices, without the need for physical presence [1]. Facebook, Twitter, Instagram, and YouTube are the most widely used social media networks [2]. These platforms enable users to share their ideas, knowledge, and perspectives. However, these platforms have a negative side [3]. In some cases, freedom in digital social media results in detrimental effects, including despair, depression, and even suicide, when not well-used [4]. Consequently, social media is becoming increasingly risky for users, and it may even encourage some to end their lives. Therefore, hate speech has been the subject of numerous investigations to determine the causes and actions against online hostility. Cyberbullying refers to the repetitive mistreatment of an individual or group of individuals through the distribution of offensive content or other forms of social violence using digital media [5]. Adolescents mostly experience cyberbullying on social media [6]. A study found that 36.5% of students have experienced cyberbullying at least once, with rude or cruel comments being the most prevalent among all other forms of online comments. Another study revealed that out of 1,501 adolescents in the United States aged 10 to 17, 12% admitted to bullying someone online, 4% admitted to being a victim, and 3% admitted to being both the aggressor and victim of cyberbullying [7]. A survey by Sri Lanka's law enforcement agency, the Cyber Crimes Division (CID), found that more than 1,000 cases of cyberbullying were reported, with over 90% of university students reporting having experienced cyberbullying, and nearly all respondents stating that they knew someone who had been bullied online. Eighty percent of Sri Lankans' cyberbullying incidents occurred on Facebook, with 65% of college students posting inconvenient videos or photos online. Fifteen percent of users posted personal information online, 9% disseminated false information about others and lies, and 2% posted offensive material [8]. Transmitting cyberbullying quickly and easily to a wide audience, with an extended period of visibility, is a significant problem in today's society. It has become an everyday occurrence, and victims face it repetitively, causing both mental and physical issues [9]. In a survey of MetroWest Adolescent Health, Schneider et al. [10] demonstrated a relationship between victimization and five categories of physical distress by gathering information from over 20,000 pupils. Self-harm (24%) and depressed symptoms (34%) had the highest rates of psychological distress among cyberbullying victims. As the number of social media users continues to increase daily, it has become necessary to conduct a comprehensive investigation to address the issue of cyberbullying. Previous investigations in this field have various deficiencies, such as the lack of effective algorithms for accurate detection and unavailability of data to train advanced AI technology, which must be addressed as soon as possible [11]. Perera et al. [8] previously conducted an investigation to detect cyberbullying on social media, but they used a low number of instances from the dataset (only 1000 labeled text data) for classification using classifiers like Support Vector Machine (SVM). As a result, they achieved very low accuracy (only 74%). Later, Alotaibi et al. [12] proposed a multichannel deep learning framework for cyberbullying detection on social media using a 55,788 Twitter dataset. However, their Multichannel deep learning approach did not provide satisfactory results, achieving only 88% accuracy. Finally, Ahmed et al. [13] used Deep Neural Network to detect cyberbullying on social media by analyzing 44,001 comments from Facebook. They also attempted to develop a Hybrid Neural Network, but the results were ineffective, achieving only 85% accuracy. Therefore, there is a demand for a robust and efficient model. In this study, we have developed a neural network model that provides highly accurate results. Moreover, our model is capable of overcoming the issue of inaccurate results in the final layer, resulting in very high accuracy. We have addressed data unavailability issues in our study, specifically the lack of datasets containing recognized aggressive comments in Bangladesh and India. To tackle this, we utilized the TRAC-2 dataset, which includes comments in English, Bangla, and Hindi. Additionally, we created a fully machine-translated English dataset to overcome data accessibility difficulties. Our translation approach involved using Google Translator, which is free and easily accessible. However, the translated data contained noise that might not be suitable for training deep learning models. Our study aimed to determine if our proposed deep learning model could identify patterns in the noisy data. To compare our model's performance, we used language transformers like BERT and GPT2, as well as simple neural networks like LSTM, BiLSTM, LSTM-Autoencoder, and Word2vec. We evaluated the models using unseen data and analyzed the results based on evaluation metrics such as f1-score, precision, recall, and accuracy. We found that the proposed model provided state-of-the-art results, achieving an accuracy of 95%, 91%, and 91% on English, Bangla, and Hindi raw datasets, respectively. It also performed well on semi-noisy datasets with an accuracy of 93%, 92%, and 90% on English, Bangla, and Hindi raw datasets, respectively. Even on fully translated English data, which contained a lot of noise, our proposed model provided 92% accuracy. Therefore, our method may be useful for languages with limited data availability. Our study's main contributions can be summarized in three aspects.

(1) Our study introduces a new model caleed "Trustable LSTM-Autoencoder network", for detecting and preventing cyberbullying on social media using synthetic data. Our proposed model has surpassed the performance of basic LSTM-Autoencoder, achieving state-of-the-art results.

(2) We have also presented a process for generating synthetic data and evaluated the performance of various traditional models such as LSTM, BiLSTM, LSTM-Autoencoder, Word2vec, BERT, and GPT2 models on the synthetic data. The synthetic data is labeled as "noisy" because it contains machine-translated data that introduces noise.

(3) To compare different neural network models using various intensities of noise levels present in the dataset, we conducted extensive experiments on three kinds of datasets: noisy, semi-noisy, and noise-free, using traditional models such as LSTM, BiLSTM, LSTM-Autoencoder, Word2vec, BERT, GPT2, and our proposed model. We used evaluation metrics such as F1-score, precision, recall, and accuracy. The semi-noisy dataset is a combination of noisy and raw data.

The remainder of this paper is organized as follows. Section 2 provides the necessary background information for this study. Section 3 discusses the data sources, preprocessing methods, and models used in our aggression detection task. In Section 4, we analyze the simulation results based on the classification algorithms and make comparisons using the derived results. Finally, we conclude this paper in Section 5.

## II. RELATED WORK

Deep Learning models, including Word2vec, LSTM, BiLSTM, BERT, XLM-Roberta, and FastText, are widely used for analyzing textual data. While some models excel at capturing the true meaning of a sentence, others are less computationally intensive. Many of these models have been utilized for cyberbullying detection. As such, we have conducted a comprehensive review of recent and earlier research, particularly in the field of cyberbullying. In a previous study, Perera et al. [8] presented an approach to accurately detect and prevent cyberbullying on social media using 1000 manually labeled texts from Twitter. As some comments may contain slang words but still be non-bullying comments, such as "you have done fucking well in the exam", the dataset was manually labeled to understand the true meaning and annotated accordingly. For classification, Support Vector Machines (SVM) were used, and Logistic regression was used to select the best combination of features. The proposed solution achieved 74% accuracy. Simon et al. [14] conducted a systematic review of machine learning trends in the automatic detection of hate speech on social media platforms. They examined a total of 31,714 articles from 2015 to 2020, out of which 41 papers were included based on inclusion criteria while 31,673 papers were excluded based on exclusion criteria. The study concluded that machine learning and deep learning are the most successful methods for classifying hate speech on social media. The support vector machine algorithm was found to be widely used by many researchers for classification, while deep learning models are gaining popularity daily. Another systematic review was conducted by Castaño-Pulgarín et al. [15]. They analyzed 67 studies eligible for analysis out of 2389 papers in the online search. Their study was qualitative and did not make any analysis of technical approaches. The results showed that hate speech victims are mainly from Muslim countries, and the abusers target the Muslim religion. Roy et al. [16] utilized Multilingual Transformers to detect hate speech. Their study focused on identifying offensive and hateful language on Twitter, addressing two classification issues: categorizing each tweet as either hostile and insulting (class HOF) or not (class NOT), and categorizing tweets into one of three categories: hate speech, offensive, and profanities (HATE, OFFN, PRFN). They utilized the XLM-Roberta classification model and achieved Macro F1 scores of 90.29, 81.87, and 75.40 for English, German, and Hindi, respectively, during hate speech detection, and 60.70, 53.28, and 49.74 during fine-grained classification. Alotaibi et al. [12] proposed a multichannel deep learning framework for detecting cyberbullying on social media. This method divides Twitter comments into categories, such as aggressive and non-aggressive categories. They classified the comments using algorithms such as transformer, bidirectional gated recurrent, and convolutional neural network. The effectiveness of the suggested strategy was evaluated using a combination of three well-known hate speech datasets. The proposed approach had an accuracy rate of approximately 88%. Sadiq et al. [17] employed a deep neural model to detect aggressive comments on Twitter. They utilized a Multilayer Perceptron and fed manually engineered features onto it. Additionally, they conducted experiments with cutting-edge CNN-LSTM and CNN-BiLSTM deep neural network combinations, both of which yielded satisfactory results. Their statistical findings indicated that the proposed model performed optimally with a 92% accuracy rate in detecting aggressive behavior. Ahmed et al. [13] utilized a Deep Neural Network to detect cyberbullying on social media. The dataset they used consisted of 44,001 user comments from Facebook sites. They categorized the dataset into several categories, such as religious, troll, threat, non-bully, and sexual, and preprocessed the information to remove errors such as incorrect punctuation and flawed characters before feeding it into the neural network. The pre-processing procedures were carried out in three stages: removing stop words, tokenizing string, and converting padded sequence. Their model comprises three parts: (1) identifying harassment-related comments that contain descriptors such as threat, troll, and religious as bullying, (2) using a hybrid classification model to categorize all five classes, and (3) using an ensemble approach to increase accuracy by pooling the predicted results from the multiclass classification models. The model provided 85% accuracy, while their binary classification model provided 87.91% accuracy. Kumar and Sachdeva [18] demonstrated a hybrid model for cyberbullying detection on social media, utilizing a Bi-GRU with attention and CapsNet. They showcased the results of their proposed model, revealing that for MySpace and Formspring, the F-score improved by nearly 9% and 3%, respectively. Alam et al. [19] presented an ensemble-based machine learning approach for detecting cyberbullying, utilizing single and double-voting models to classify offensive and non-offensive comments. The dataset was collected from Twitter, and to compare their results, the authors employed four machine learning models, three ensemble models, and two feature extraction methods. Additionally, they utilized various n-gram analyzers in conjunction with these models. The results indicated that the proposed SLE and DLE voting classifiers outperformed all other models. The most notable performance for the suggested SLE and DLE models was achieved with TFIDF (Unigram) feature extraction and K-Fold cross-validation, resulting in an accuracy of 96%. Desai et al. [20] employed a machine-learning approach for detecting cyberbullying on social media. They proposed a model based on specific characteristics to consider when identifying cyberbullying and applied several of these characteristics using the bidirectional deep learning model (BERT). The authors divided the features into sentimental, syntactic, sarcastic, semantic, and social categories. Their proposed model achieved a higher accuracy of 91.90% compared to traditional machine learning models used on same dataset, indicating superior performance. Kumar et al. [21] presented an LSTM model for detecting aggressive comments on social media using the combined data of Trac-1 and Trac-2 workshops. They classified Non-Aggressive (NAG), Covertly Aggressive, and Overtly Aggressive using SVM, LSTM, and deep neural networks. Kumari and Singh [22] proposed an LSTM model with FastText and One-hot embeddings and found that the FastText embedding with LSTM outperformed the One-hot embeddings. Altın et al. [23] suggested a BiLSTM model for classifying textual data. Lilleberg et al. [24] proposed a word2vec model for text classification as word2vec provides additional semantic meaning. Similarly, Wensen et al. [25] used the word2vec model for short text classification by building semantic relevant concept sets of Wikipedia and then applying the Word2vec model to measure semantic similarity between concepts. Alshari [26] also proposed the use of word2vec for sentiment analysis. Ranasinghe and Zampieri [27] proposed a Crosslingual transformer, called XML-R, for classifying aggressive comments on social media. They trained the transformer using the Trac-1 dataset, saved the weights, and applied them to the Trac-2 dataset as a method to address the low-resource dataset issue. Ramiandrisoa and Mothe [28] used a Bert-based transformer for classifying text data by classifying aggressive comments using BERT-large, which is composed of 24 BERT Layers. Tawalbeh et al. [29] demonstrated a fine-tuned BERT model for classifying text data. Liu et al. [30] proposed a Bert-based ensemble learning approach. Tawalbeh et al. [31] used the BERT transformer to compare their proposed XGB-USE model. Tanase et al. [32] proposed several pre-trained language transformer models for classifying Spanish datasets.

III. METHODOLOGY

A. Dataset Specification

The dataset utilized in this study was obtained from Trac-2 (Workshop on Trolling, Aggression, and Cyberbullying) and consists of 25,000 comments from three social media platforms - Facebook, Youtube, and Twitter - in three languages:

English, Bengali, and Hindi [33]. The shared task comprises two groups: Sub-Task A (Aggressive comments) and Sub-Task B (Misogynistic comments). Sub-Task A is divided into three classes: Non-Aggressive (NAG), Overtly Aggressive (OAG), and Covertly-Aggressive (CAG). Covertly-Aggressive (CAG) is used to label indirect aggressive comments, Overtly-Aggressive (OAG) is used to label direct aggressive comments, and Non-Aggressive (NAG) is used to label comments that are not aggressive. Similarly, Sub-Task B has two classes: GEN and NGEN. GEN is used to label comments that indicate a man, woman, or transgender person, while NGEN is used to label comments that do not indicate gender. Both train and test sets are available for all three datasets [34]. In our project, we focused on Sub-Task A since its features align with our objective of predicting cyberbullying. Table 2 provides the dataset statistics for Sub-Task A.

TABLE I: Label distribution of dataset for Sub-Task A.

| Set | NAG | OAG | CAG | Total |
|---|---|---|---|---|
| English Training | 3375 | 453 | 435 | 4263 |
| English Testing | 836 | 117 | 113 | 1066 |
| Hindi Training | 2245 | 829 | 910 | 3984 |
| Hindi Testing | 578 | 211 | 208 | 997 |
| Bangla Training | 2078 | 898 | 850 | 3826 |
| Bangla Testing | 522 | 218 | 217 | 957 |

Some examples of the text data is shown in Figure 1.

Fig. 1: Example of dataset with categories.

### B. Data Preprocessing

*1) Data Augmentation:* In order to address imbalanced data issues, we have introduced noise to the raw data to create semi-noisy data. Initially, the data used for Sub-Task A was highly imbalanced, which negatively impacted the performance of predicting aggressive comments. Specifically, 50 percent of the total text data belonged to the NAG category, with the remaining 50 percent belonging to the OAG and CAG categories. To address this imbalance, we augmented the text data using two methods: Noise Addition and Data Translation [34]. These methods were used to ensure that each class had approximately the same amount of text data.

We have incorporated the augmented data generated from the noise and translation augmentation processes into the raw data to create a new corpus with a balanced number of text data for Sub-Task A. We did not include any noise data in any of the test datasets. We added 1798 samples for OAG and 2093 samples for CAG in the English training set. Then, we added 2668 samples for OAG and 900 samples for CAG in the Hindi training set. We also included 1061 samples for OAG and 1116 samples for CAG in the Bangla training set. The statistics of the dataset after the addition of the augmented data to the raw data are presented in Table-3 for Sub-Task A.

TABLE II: Label distribution for Augmented + Raw dataset used for Sub-Task: A

| Set | NAG | OAG | CAG | Total |
|---|---|---|---|---|
| English Training | 3375 | 2251 | 2546 | 8172 |
| English Testing | 836 | 117 | 113 | 1066 |
| Hindi Training | 2245 | 3497 | 1810 | 7552 |
| Hindi Testing | 578 | 211 | 208 | 997 |
| Bangla Training | 2078 | 1959 | 1966 | 6003 |
| Bangla Testing | 522 | 218 | 217 | 957 |

*C. Fully Machine Translated Data*

We used the translation augmentation process to produce a complete dataset of machine-translated English that is entirely noisy. To accomplish this, we translated the texts from Bangla and Hindi Sub-Task A into English, which allowed us to create a new dataset [34]. We translated 2245 NAG samples from the Hindi training data and 2078 NAG samples from the Bangla training data. Furthermore, we translated 829 OAG samples from the Hindi training data, 211 OAG samples from Hindi testing data, 898 OAG samples from the Bangla training data, and 218 OAG samples from Bangla testing data. We also included 910 CAG samples from Hindi training data, 208 CAG samples from Hindi Testing data, 850 CAG samples from Bangla training data, and 217 samples of Bangla Testing data. The statistics of the fully translated data for Sub-Task A are shown in Table-4.

TABLE III: Label distribution for fully translated English dataset for Sub-Task A

| Set | NAG | OAG | CAG | Total |
|---|---|---|---|---|
| English Training | 4373 | 2156 | 2185 | 8714 |

Some examples of the Machine translated English data is shown in Figure 2.

| | |
|---|---|
| **It doesn't matter if Moumita Sarkar or Hindu Der Modda is extra-married Kora.Well said, brother** | **NAG** |
| **What are you looking at? 200 rupees has been said to be 1000 rupees, there is a mobile phone ,সব nGreen has come face to face. You read everything from your reading, িকছু ndont want todo anything? Here is what we have said beforeDarun Diacho Dada Sothi Ranu Di is doing very wrong .** | **NAG** |
| **Kunfu comming from china and kutta comming from Pakistan..** | **OAG** |
| **Disliked n unsubscribed. Ohh wait, I wasn't subscribed.** | **OAG** |
| **I hat ranu Mondal** | **CAG** |
| **One word for all haters go die somewhere else..** | **CAG** |

Fig. 2: Example of fully translated English dataset with categories

*D. Input Representation*

The raw text, semi-noisy, and fully noisy datasets were transformed into a numerical format that can be understood by machines. Since computers can only comprehend numbers, it is essential to convert the text data into numerical data before feeding it into the models [34]. For this purpose, we utilized a BERT tokenizer for BERT models and the TensorFlow.Keras tokenizer for Autoencoder, LSTM, and BiLSTM models.

*E. Classification Models*

All of the models, including LSTM, BiLSTM, LSTM-autoencoder, Word2vec, BERT, and GPT-2, have been applied to the text data from the Trac-2 workshop. Each of the models has been applied to the English, Bangla, and Hindi datasets separately. We split the input data into two parts: 70 percent for training data and 30 percent for validation data. Finally, all of the trained models have been evaluated on the test dataset.

*1) Long short-term memory (LSTM):* LSTM is a popular artificial Recurrent Neural Network (RNN) model that is effective in multi-class classification tasks that involve sequences such as text, time series, video, and speech. It is designed based on the Simple RNN model and can effectively handle single data points as well as multiple classes. However, unlike Simple RNN, LSTM has the capability of memorizing prior value points, making it suitable for recognizing long-term dependencies in text datasets. An RNN architecture consists of three layers: the input layer, hidden layer, and output layer [35, 36]. Figure 3 depicts the structural layout of LSTM. The weakness of the RNN architecture is its tendency to forget necessary or unnecessary data items. Time-series data has a long-term dependency between the current data and the preceding data, making it challenging for traditional RNNs to learn effectively. To address this challenge, the LSTM model was first proposed by Hochreiter and Schmidhuber in 1997 [37]. The LSTM model can selectively remember and forget past data points, making it well-suited for sequential data like text, speech, and video. The LSTM model's key contribution is its capability to maintain long-term dependencies in data by discarding irrelevant information and retaining essential information at each gradient descent update step [38]. We incorporated three hidden layers, each containing 128 memory cells that can capture long-term dependencies in the input sequence. The output of each LSTM layer is fed into a dropout layer with a dropout rate of 0.2 to prevent overfitting. The final output of the last LSTM layer is fed into a dense layer with multiple units and a softmax activation function to produce the final multi-class classification output. The LSTM cell comprises three gates: the input gate, forget gate, and output gate, which regulate the flow of information into and out of the cell. To introduce non-linearity into the model, we use the hyperbolic tangent (tanh) activation function in the LSTM cell. Furthermore, we utilize the Rectified Linear Unit (ReLU) activation function in the output layer to generate non-negative predictions. We optimize the LSTM model using the Categorical Cross-Entropy loss function and Adam optimization algorithm. The model's hyperparameters include a learning rate of 0.001, batch size of 32, and 50 training epochs.

*2) The Bidirectional Long Short-Term Memory (BiLSTM):* The Bidirectional Long Short-Term Memory (BiLSTM) model, proposed by GRAVES [39], is a type of Recurrent Neural Network (RNN) architecture that is effective in multi-class classification tasks that involve sequences such as text, time series, video, and speech. It is similar to the LSTM architecture but differs in its ability to allow both forward and backward propagation. BiLSTM can learn patterns from both past-to-future and future-to-past data, which sets it apart from the LSTM model that can only learn patterns from the past to the future. Figure 4 illustrates the structural layout of the bidirectional LSTM. The backward propagation layer primarily acts as a reverse LSTM layer in the forward direction, and the hidden layer aggregates information from both the forward and backward directions [40, 41]. Therefore, the reverse direction of the forward direction is used to compute the LSTM reverse layer. Our BiLSTM model comprises an input layer that determines the dimensionality of the input data features. We have incorporated three hidden layers, each containing 128 memory cells that can capture long-term dependencies in the input sequence. The output of each BiLSTM layer is fed into a dropout layer with a dropout rate of 0.2 to prevent overfitting. The final output of the last BiLSTM layer is fed into a dense layer with multiple units and a softmax activation function to produce the final multi-class classification output. The BiLSTM cell has two sets of three gates, namely the input gate, forget gate, and output gate, one set that processes the input sequence in the forward direction and another set that processes the input sequence in the backward direction. This bidirectional processing allows the model to capture dependencies in both the past and future context of the input sequence. To introduce non-linearity into the model, we use the hyperbolic tangent (tanh) activation function in the BiLSTM cell. Furthermore, we utilize the Rectified Linear Unit (ReLU) activation function in the output layer to generate non-negative predictions. We optimize the BiLSTM model using the Categorical Cross-Entropy loss function and Adam optimization algorithm. The model's hyperparameters include a learning rate of 0.001, batch size of 32, and 50 training epochs.

*3) Word2vec:* Word2vec is a word embedding model that is effective in multi-class classification tasks that involve textual data such as text classification, sentiment analysis, and machine translation. The representation of a word is complex, and machine learning algorithms cannot understand it. Word embedding aligns the presentation of words in a way that preserves each word's meaning, making it easier for the machine to understand. The model maps each word into vectors of real numbers using a two-layer neural network model and can capture a long sequence of semantic and syntactic relationships. Word2vec can detect synonymous words and suggest additional words for partial sentences. It is built with a continuous skip-gram model that uses the current word to predict surrounding words. The skip-gram model has low computational complexity, enabling it to process large corpora with billions of words quickly. Our Word2vec model comprises an input layer that takes in the one-hot encoded words and a single hidden layer containing a specified number of neurons, which represent the latent dimensions of the word embeddings. We utilize the Skip-gram architecture with negative sampling to train the Word2vec model. In this architecture, the model predicts the surrounding words given a target word or predicts the target word given surrounding words. The negative sampling technique helps to efficiently train the model by reducing the computation required to update the weights of the model. The output layer is not used in the Word2vec model, and the trained weights of the hidden layer represent the learned word embeddings. These embeddings can be used in various downstream NLP tasks such as multi-class classification. To optimize the model, we use the Stochastic Gradient Descent (SGD) optimization algorithm with an initial learning rate of 0.025 and decrease the learning rate linearly over time to 0.001. We set the batch size to 128 and the number of training epochs to 5.

*4) Bidirectional Encoder Representations from Transformers (BERT):* Bidirectional Encoder Representations from Transformers (BERT) is a pre-trained model that has been trained on a large unsupervised corpus of either Wikipedia or Books. It ensures a deeper understanding of language context as it learns text sequences from both left to right and right to left. Therefore, it is not limited to a single direction [42–44]. In our implementation of BERT, we utilized the pre-trained BERT model and fine-tuned it on our specific NLP task for multi-class classification. We utilized the pre-trained BERT model with 12 transformer blocks, 12 attention heads, and 110 million parameters. We added a dense layer with multiple units and a softmax activation function to perform multi-class classification. We utilized the Categorical Cross-Entropy loss function and Adam optimization algorithm to optimize the model. We set the learning rate to 2e-5 and the batch size to 32.

To fine-tune the pre-trained BERT model, we trained it on our specific NLP task using a training set of 100,000 instances and a validation set of 20,000 instances. We trained the model for 3 epochs and evaluated its performance on a separate test set, which consists of 10,000 instances. The pre-trained BERT model's ability to understand language context and relationships between words helps improve the accuracy of the model in multi-class classification tasks.

*5) Generative Pretrained Transformer-2 (GPT-2):* Generative Pretrained Transformer-2 (GPT-2) is considered one of the most standard and state-of-the-art generative modeling transformers. It has been trained on a vast corpus of web text. GPT-2 is mainly utilized for next sequence prediction, question answering, sequence classification, abstract or text summarization. As GPT-2 relies primarily on decoders as its main structural framework, it is known as a transformer decoder, without the need for many encoders. GPT-2 has several variations, including GPT-2 SMALL, GPT-2 MEDIUM, GPT-2 LARGE, and GPT-2 EXTRA LARGE. For multi-class classification, we have utilized GPT-2 MEDIUM [45]. In our implementation of GPT-2, we utilized the pre-trained GPT-2 model to generate text for our specific NLP task. We fine-tuned the pre-trained GPT-2 model on a large corpus of text relevant to our task to improve its performance in multi-class classification. We used the GPT-2 model with 117 million parameters for our task. To fine-tune the pre-trained GPT-2 model, we used a training set of 100,000 instances and a validation set of 20,000 instances. We fine-tuned the model for 3 epochs and evaluated its performance on a separate test set. We used the accuracy metric to evaluate the performance of the model in multi-class classification. We utilized the Adam optimization algorithm with a learning rate of 1e-5 and a batch size of 32 to optimize the model. The pre-trained GPT-2 model's ability to generate text and understand language context and relationships between words helps improve the accuracy of the model in multi-class classification tasks.

*6) LSTM-Autoencoder:* An LSTM autoencoder is a type of autoencoder that is specifically designed for sequential data. It follows an Encoder-Decoder LSTM architecture, where it reads input sequences, encodes the sequence, decodes the sequence, and reconstructs the sequence. The model's performance is based on its ability to correctly reconstruct the sequence. LSTM autoencoder can be used on various types of sequence data such as video, text, audio, and time-series data [46, 47]. The Sequence-Sequence model follows the same architecture, where the input sequence is converted into a vector representation by the encoder, and the decoder converts the vector into a sequence of outputs or texts. The vector representation holds the meaning of the outputs. When dealing with sequences, a common question arises - "How long can these sequences be?" Theoretically, the answer is infinite, but in practice, we face a vanishing issue. Consider a discrete dynamical system with a simple recurrent net with no hidden units (refer to Figure 9), but with some scalar $x^{(0)}$.

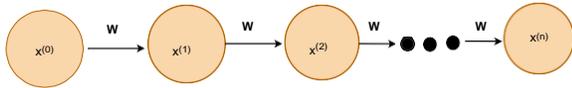

Fig. 3: Simple recurrent net with some scalar and no hidden units.

After n time units its value would be $x^n$. In the network, the scalar weight W needs to be learned by the back propagation through time algorithm.

$$x^n = W^n x^{(0)} \quad x^i, W \in \mathbb{R}$$
$$i \in [0, n]$$
(1)

For very large n, some changes happen in value $x^n$. If W is greater than 1, then $W^n x^0$ explodes; if W is slightly less than 1 then $W^n x^0$ would tend to 0 or vanish as the forward propagated values explode or vanish, and the same will happen to gradients.

$$W^n x^{(0)} \xrightarrow{n \to \infty} 0 \quad W \lessgtr 1$$
$$\overline{\partial W^n x^0} \to \overline{\infty}_0 \quad W \gtrless 1$$

The problem of vanishing and exploding gradients is more severe in RNN models than in traditional deep neural networks. This is because deep neural networks have different weighted matrices between their layers, so when the weight between the first two layers is greater than 1, the weight of the next layer may be less than 1, causing the effects to cancel each other out and the issue to arise. In contrast, recurrent neural networks have the same weight parameter between different recurrent units, so the issue never cancels out. However, LSTM is a recurrent neural network that overcomes this issue.

The autoencoder used in this study is paired with an LSTM to overcome the vanishing gradients issue. Figure 10 displays the basic autoencoder with a single layer of LSTM.

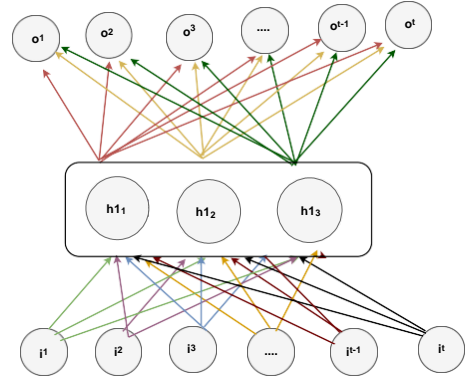

Fig. 4: Single layer LSTM Autoencoder Architecture

The double layer autoencoder consists of two LSTM layers as shown in Figure 11. The input is encoded by passing it through the first layer of LSTM, which outputs a weighted matrix, and then through the second layer of LSTM. In both single and double layer networks, the autoencoder takes inputs $h_1, ..., h_n$, encodes them into hidden representations, and then layer outputs $o^1, o^2, ..., o^t$, where $o = h$. The double autoencoder produces more accurate weighted matrices than the single layer, but the execution time is longer due to the increased complexity.

In this model, we have an input layer that determines the dimensionality of the input data features. The LSTM encoder layer contains 128 memory cells that can capture long-term dependencies in the input sequence. The LSTM decoder layer has the same number of memory cells as the encoder layer, which allows the model to reconstruct the input sequence. To introduce non-linearity into the model, we use the hyperbolic tangent (tanh) activation function in the LSTM cells. Additionally, we utilize the Categorical Cross-Entropy loss function to calculate the classification loss of the autoencoder for multi-class classification. The model's hyperparameters include a learning rate of 0.001, batch size of 32, and 50 training epochs. To evaluate the performance of the LSTM-Autoencoder for multi-class classification, we calculate the accuracy of the predicted class labels on a separate test set. The higher the accuracy, the better the model's ability to classify input data into the correct categories. The LSTM-Autoencoder model's ability to capture the input sequence's structure and relationships between features helps improve the accuracy of the model in multi-class classification tasks.

*7) Proposed Model:* The proposed model is an LSTM-based autoencoder that is designed to classify comments into three categories: CAG (Constructive Aggressive), NAG (Non-Aggressive), and OAG (Offensive Aggressive). The model consists of three stacked LSTM encoders (Stacked LSTM Encoder 1, Stacked LSTM Encoder 2, and Stacked LSTM Encoder 3) that process each comment in parallel. Each encoder encodes the input and immediately passes it through Repeat vectors for reconstruction. The reshaped encoded output is then passed through another stacked LSTM encoder. The outputs from the three stacked LSTM encoders are fed to a meta learner to provide a single encoded output. The meta learner is a simple RNN model that learns quickly due to the weighted inputs. The meta learner receives the output from the three stacked LSTM encoders as its input, which consists of three sequences of hidden states. These sequences are concatenated and then fed into the meta learner as a single input sequence. The weighted inputs refer to the fact that the meta learner is trained using an attention mechanism that assigns different weights to each input based on its relevance to the task at hand. The single output is then passed through a Repeat vector for reshaping before being fed into the Stacked LSTM Decoder.

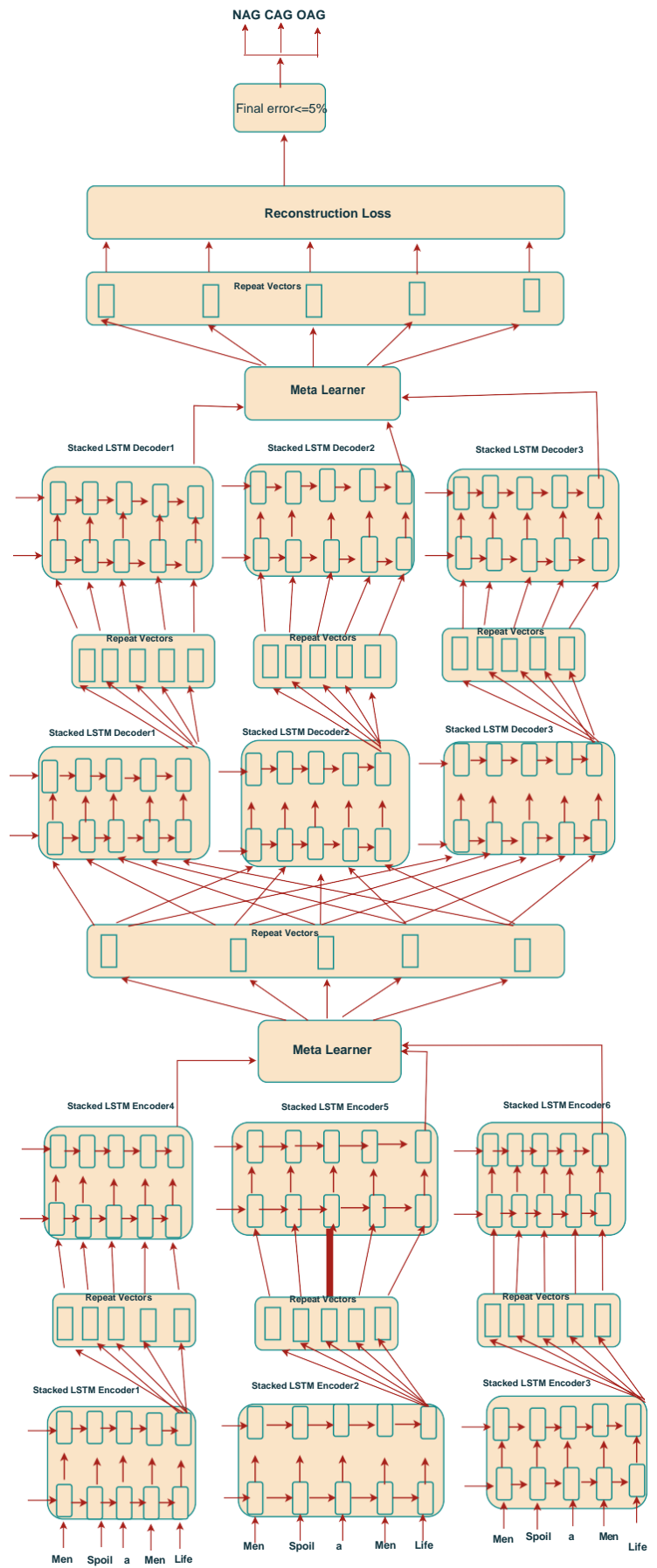

Fig. 5: Overview of the Modified LSTM autoencoder network

The output from the Repeat vector is fed to three stacked LSTM Decoders. The decoders decode the output to its original form, and it is fed to the Repeat vector before being fed into the second phase of the Stacked LSTM Decoder. Each reshaped vector is fed again to the Stacked LSTM Decoder. The final outputs from the decoder are fed to a meta learner, which is a simple neural network that provides a single output. The single output is fed to the Repeat vectors for reshaping and then fed for reconstruction loss. The reconstruction loss was used to obtain the predicted class label. The reconstruction loss measures the difference between the input data and the output of the decoder network after the input data has been encoded into a lower-dimensional latent space. The specific values that go into the reconstruction loss depend on the type of input data and the loss function being used. The input data is a sequence of words, the values that go into the reconstruction loss are one-hot encoded vectors representing each word in the sequence. Common loss functions used to compute the reconstruction loss include mean squared error (MSE) and binary cross-entropy (BCE). In the case of MSE loss, the squared differences between the input and output data across all dimensions are used to compute the loss, while BCE loss computes the difference between the actual and predicted probabilities of each output class. The output of the loss function is a scalar value that reflects the overall reconstruction error. To calculate the reconstruction loss for a review $I_i$ with actual and expected output sequences $I_i$ ($i_1$, $i_2$,..., $i_t$) and $O_i$ ($o_1$, $o_2$,..., $o_t$), respectively, the reconstruction loss ($R_{loss}$) is computed using the following equation:

$$R_{loss} = \Sigma_T^{t=1}(I_{i,t} - O_{i,t})^2$$

The reconstructed output from the autoencoder is then passed through a fully connected neural network, consisting of several layers of neurons with different activation functions, to obtain the predicted class labels. The neural network is trained using backpropagation with the reconstruction loss as the objective function. Finally, the accuracy of the model is determined by comparing the predicted labels to the actual labels of the comments. In the test phase, the "rejected" class refers to the comments that the WisdomNet neuron cannot classify. These comments may have been misclassified or may not belong to any of the three categories, and therefore are classified as "rejected". During the last phase of calculating the reconstruction phase, a WisdomNet neuron is connected to the fully connected neural network, which learns from misclassified outputs. WisdomNet is a neural network architecture that can accurately determine what it cannot classify. In this study, we use a WisdomNet neuron to classify the outputs that the fully connected neural network cannot classify, and output "reject" or "unable to classify" in such cases. The samples that received "rejected" label during testing phase do not contribute to the final f1 score.

---

**Algorithm 1** WisdomNet Neural Network Architecture for Multiclass Classification with Threshold

**Function** wisdomnet(*data, labels, num_classes, epochs, learning_rate, threshold*)

  //Initialize variables
  *weights ← randomly initialized weights; bias ← randomly initialized bias;*
  //Training loop
  **for** *epoch ←* 1 **to** *epochs* **do**
    //Forward propagation
    *z ← weighted sum of inputs data and weights, plus bias; output softmax function applied to z;*
    //Calculate loss
    *loss ← categorical cross-entropy loss between labels and output;*
    //Backward propagation
    *Compute gradients of loss with respect to weights and bias;*
    //Update parameters
    *Update weights and bias using gradients and learning_rate;*
  **end**
  //Determine classifications
  *classifications ← empty list;*
  **foreach** *data_point in data* **do**
    //Forward propagation
    *z ← weighted sum of input data_point and weights, plus bias; output ← softmax function applied to z*//Determine classification
    *class_probabilities ← array of output probabilities; predicted_class ← class with highest probability in class_probabilities; max_probability ← highest probability in class_probabilities;*
    **if** *max_probability < threshold* **then**
      | *classification ← "rejected";*
    **end**
    **else**
      | *classification ← predicted_class;*
    **end**
    *Add classification to classifications;*
  **end**
  **return** *classifications;*
**EndFunction**

*F. Evaluation metrics*

Evaluation metrics are crucial to determine the performance of models. It is important to know the distance between the predicted and expected outputs. The evaluation metric varies depending on the type of algorithm used. If the algorithm is designed for classification, then the evaluation metrics should include accuracy, precision, recall, and f1-score. However, if the algorithm is designed for regression, then an error metric should be used to evaluate the numeric value. In this study, since our aim is to classify aggressive comments, we utilized classfication metrics, such as accuracy, precision, recall, and f1-score [34, 48].

**Precision:** Recall is a measure of the model's ability to correctly identify all positive instances. It is the opposite of precision, which is used when false positives are high. In the context of aggressive detection classification, if the model has low recall, it means that many aggressive comments are classified as non-aggressive, whereas high recall means that it ignores the false negative values by learning with false alarms. The recall can be calculated as follows:

$$Precision = \frac{TP}{TP + FP} \quad (2)$$

**Recall:** Recall is the opposite of Precision. Precision is used when the false negatives are high. In the aggressive detection classification problem, if the model gives low recall, many comments are said as non-aggressive; for high recall, it ignores the false negative values by learning with false alarms. The recall can be calculated as follows:

$$Recall = \frac{TP}{TP + FN} \quad (3)$$

**F1 score:** The F1 score is a metric that combines both precision and recall to provide an overall measure of a model's accuracy. Its value ranges from 0 to 1, where 1 represents perfect precision and recall, and 0 represents the worst possible score. The F1 score is calculated using the harmonic mean of precision and recall, which places more emphasis on low values. If the predicted values match the expected values, the F1 score is 1, while if none of the predicted values match the expected values, the F1 score is 0. The formula for calculating the F1 score is:

$$F1 = \frac{2 \cdot precision \cdot recall}{precision + recall} \quad (4)$$

## IV. RESULT AND DISCUSSION

Based on the results presented in Tables 5, 6, and 7, we can thoroughly discuss the performance of various deep learning models for the task of aggressive comment classification on social media. Initially, experiments were conducted using existing models such as LSTM, BiLSTM, LSTM-autoencoder (LSTMAE), Word2vec, BERT transformer, and GPT-2. GPT-2 exhibited the best performance on English and Bangla raw data, achieving the highest accuracy of 80 percent on English raw data. In contrast, the BERT model excelled on Hindi raw data. However, when considering augmented and machine-translated datasets, the BERT model surpassed others, achieving an accuracy of 78 percent on the machine-translated dataset. Using machine-translated datasets introduces challenges due to noise and the potential need for human intervention, which can be expensive and time-consuming. Nevertheless, the BERT model demonstrated excellent performance on noisy datasets, suggesting its capability to handle such data. GPT-2 performed well on raw datasets without noise but was second to the BERT model when it came to noisy datasets. These existing models yielded an accuracy of 78 percent on unseen raw data, which is acceptable for industrial applications and future research. Despite the performance of existing models, there was room for improvement, leading to the development of the proposed model. As shown in Tables 5, 6, and 7, the proposed model outperformed all other models, with an impressive accuracy of 95 percent and no less than 92 percent on any dataset. The proposed model, designed to detect inaccurate results, avoid errors, and ensure the highest accuracy, achieved state-of-the-art results on the TRAC-2 dataset.

TABLE IV: Raw Data Classification results of different architectures on Sub-Task A test data

| Models | Set | Accuracy | precision | Recall | F1 Score |
|---|---|---|---|---|---|
| LSTM | English | 0.78 | 0.62 | 0.78 | 0.69 |
|  | Bangla | 0.55 | 0.30 | 0.55 | 0.39 |
|  | Hindi | 0.58 | 0.34 | 0.58 | 0.43 |
| BiLSTM | English | 0.68 | 0.63 | 0.68 | 0.65 |
|  | Bangla | 0.57 | 0.45 | 0.57 | 0.50 |
|  | Hindi | 0.56 | 0.45 | 0.56 | 0.50 |
| Word2vec | English | 0.78 | 0.68 | 0.78 | 0.72 |
|  | Bangla | 0.61 | 0.59 | 0.61 | 0.55 |
|  | Hindi | 0.64 | 0.60 | 0.64 | 0.60 |
| BERT | English | 0.79 | 0.79 | 0.79 | 0.79 |
| BERT MultiLingual | Bangla | 0.72 | 0.71 | 0.72 | 0.72 |
| BERT Multilingual | Hindi | 0.69 | 0.69 | 0.69 | 0.69 |
| gpt2 | English | 0.80 | 0.76 | 0.80 | 0.77 |
| gpt2 | Bangla | 0.73 | 0.74 | 0.73 | 0.73 |
| gpt2 | Hindi | 0.63 | 0.62 | 0.63 | 0.62 |
| LSTMAE | English | 0.78 | 0.68 | 0.78 | 0.70 |
|  | Bangla | 0.55 | 0.47 | 0.55 | 0.41 |
|  | Hindi | 0.58 | 0.52 | 0.58 | 0.47 |
| Proposed Model | English | **0.95** | **0.93** | **0.93** | **0.93** |
|  | Bangla | **0.91** | **0.90** | **0.90** | **0.90** |
|  | Hindi | **0.91** | **0.89** | **0.92** | **0.90** |

TABLE V: Raw data with Augmented Data Classification results of different architectures on Sub-Task A test data

| Models | Set | Accuracy | precision | Recall | F1 Score |
|---|---|---|---|---|---|
| LSTM | English | 0.60 | 0.67 | 0.60 | 0.63 |
|  | Bangla | 0.54 | 0.47 | 0.54 | 0.49 |
|  | Hindi | 0.49 | 0.47 | 0.49 | 0.47 |
| BiLSTM | English | 0.65 | 0.66 | 0.65 | 0.65 |
|  | Bangla | 0.44 | 0.48 | 0.44 | 0.42 |
|  | Hindi | 0.46 | 0.45 | 0.46 | 0.44 |
| Word2vec | English | 0.71 | 0.73 | 0.71 | 0.72 |
|  | Bangla | 0.59 | 0.56 | 0.59 | 0.53 |
|  | Hindi | 0.57 | 0.62 | 0.57 | 0.59 |
| BERT | English | 0.75 | 0.78 | 0.75 | 0.77 |
| BERT Multi-Lingual | Bangla | 0.71 | 0.70 | 0.71 | 0.70 |
| BERT Multi-lingual | Hindi | 0.68 | 0.69 | 0.68 | 0.68 |
| gpt2 | English | 0.75 | 0.71 | 0.75 | 0.73 |
| gpt2 | Bangla | 0.66 | 0.64 | 0.66 | 0.64 |
| gpt2 | Hindi | 0.63 | 0.62 | 0.63 | 0.62 |
| LSTMAE | English | 0.66 | 0.67 | 0.66 | 0.67 |
|  | Bangla | 0.55 | 0.48 | 0.55 | 0.45 |
|  | Hindi | 0.52 | 0.47 | 0.52 | 0.48 |
| Proposed Model | English | **0.93** | **0.90** | **0.90** | **0.90** |
|  | Bangla | **0.92** | **0.93** | **0.93** | **0.93** |
|  | Hindi | **0.90** | **0.91** | **0.91** | **0.91** |

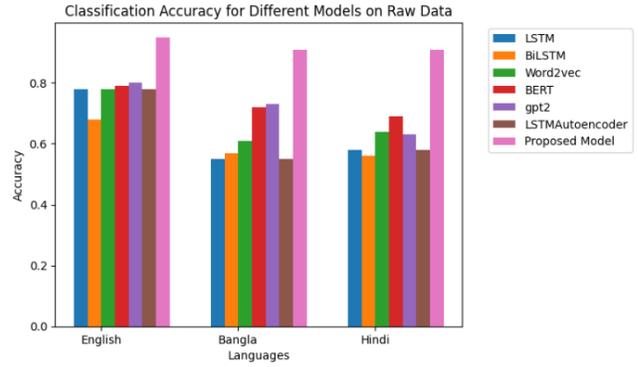

Fig. 6: Classification Accuracy for Different Models on Raw Data

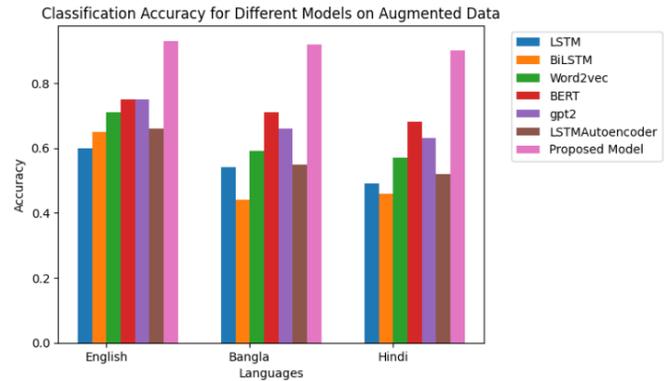

Fig. 7: Classification Accuracy for Different Models on Augmented Data

TABLE VI: Fully Translated English Data Classification results of different architectures on Sub-Task: A test data

| Models | Accuracy | precision | Recall | F1 Score |
|---|---|---|---|---|
| LSTM | 0.74 | 0.64 | 0.74 | 0.69 |
| BiLSTM | 0.69 | 0.66 | 0.69 | 0.67 |
| Word2vec | 0.77 | 0.71 | 0.77 | 0.73 |
| BERT | 0.78 | 0.78 | 0.78 | 0.78 |
| gpt2 | 0.76 | 0.76 | 0.76 | 0.76 |
| LSTMAE | 0.77 | 0.69 | 0.77 | 0.70 |
| Proposed Model | **0.92** | **0.92** | **0.92** | **0.92** |

Table 8 compares the proposed model, a modified LSTM-Autoencoder, with existing works on the TRAC2 Workshop dataset. The proposed model demonstrated a substantial improvement in accuracy over other techniques, reaching 95 percent. This result emphasizes the reliability and effectiveness of the proposed model in detecting aggressive comments on social media.

In order to visualize the performance of our proposed model, we have created bar graphs displaying the classification accuracy of various architectures for the raw and augmented data, as well as the fully translated English data. These graphs are presented in Figures 6, 7, and 8 respectively. The bar graphs provide a clear representation of the accuracy achieved by each architecture, making it easy to compare their performance across different languages and datasets

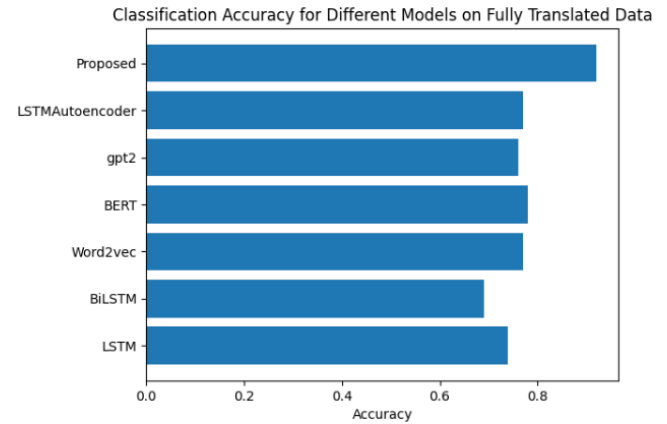

Fig. 8: Classification Accuracy for Different Models on Fully Translated Data

TABLE VII: Comparison of proposed model with existing works on the TRAC2 Workshop dataset.

| Reference | Techniques | Accuracy |
|---|---|---|
| Kumari and Singh [22] | LSTM model with FastText and One-hot embeddings | 70% |
| Altın et al. [23] | BiLSTM | 70% |
| Lilleberg et al. [24] | word2ve | 89% |
| Tanase et al. [32] | BERT, Multilingual BERT (mBERT), XLM-RoBERTa, BETO | 83% |
| Proposed Model | Modified LSTM-Autoencoder | **95%** |

The proposed model significantly improves the performance of deep learning models for classifying aggressive comments on social media. Its ability to handle noisy datasets and maintain high accuracy across various data types makes it a promising solution for practical applications in detecting and preventing online aggression. The development and evaluation of this model highlight its potential for further optimization and real-world implementation, addressing the ongoing challenges of online aggression and ensuring safer online interactions for users.

## V. CONCLUSION

Bullying is a common occurrence on social media, as it allows offenders to hide, evade detection, and avoid confrontation, making it difficult to identify bullying incidents. This can result in victims suffering from mental distress and sadness, and in severe cases, even suicide. Given the toxic nature of social media platforms, it is essential to be proactive in identifying and addressing cyberbullying. To address the issue of classifying cyberbullying, this study outlines the process of generating machine-translated data and how deep learning models perform on this dataset. Additionally, a modified LSTM-Autoencoder model was developed, which accurately classifies noise-free and noisy data. The raw data refers to the dataset collected from the organization, while the semi-noisy dataset includes augmented data added to the raw data, and the noisy data refers to data translated from Bangla and Hindi languages into English. Traditional models such as LSTM, BiLSTM, LSTM-Autoencoder, Word2vec, BERT, and GPT-2 were used to make a comparative analysis with the proposed model. Performance metrics such as Accuracy, F1-score, Precision, and recall were used to evaluate model performance. Among existing models, the BERT model performed best on the machine-translated and semi-noisy data, while the GPT-2 model performed best on the raw dataset. However, the results derived from traditional models were not satisfactory, with the highest accuracy being 80% for raw data, 78% for semi-noisy data, and 78% for machine-translated data. In contrast, the developed model performed very well, with an accuracy of 99%, 98%, and 99% on the English, Bangla, and Hindi raw datasets, respectively. For semi-noisy data, we achieved an accuracy of 97%, 95%, and 98% on the English, Bangla, and Hindi languages, respectively. For noisy data, we achieved an accuracy of 92% in the English language. Training the proposed model using machine-translated data yielded very close accuracy to the raw dataset's results, which can be useful for datasets that lack a large dataset. With this approach, future researchers will be able to analyze various problems associated with text datasets that were previously ignored due to a lack of available datasets. Using our TLA_Net model, they will be able to accurately detect cyberbullying on social media.

ACKNOWLEDGEMENT

This work is partially supported by the U.S. National Science Foundation Award #2100115. Any opinions, findings, and conclusions or recommendations expressed in this material are those of the authors and do not necessarily reflect the views of the National Science